\documentclass[conference]{IEEEtran}
\IEEEoverridecommandlockouts
\usepackage{cite}
\usepackage{amsmath,amssymb,amsfonts}
\usepackage{algorithmic}
\usepackage{graphicx}
\usepackage{textcomp}
\usepackage{xcolor}
\usepackage{bbding}
\usepackage{amsmath}
\usepackage{color}
\usepackage{bbm}
\usepackage{multirow}
\usepackage{makecell}
\usepackage{caption}
\usepackage{subcaption}
\usepackage{enumitem}

\def\BibTeX{{\rm B\kern-.05em{\sc i\kern-.025em b}\kern-.08em
    T\kern-.1667em\lower.7ex\hbox{E}\kern-.125emX}}
\begin{document}

\title{VIAssist: Adapting Multi-modal Large Language Models for Users with Visual Impairments  
% \\
}

% \author{Bufang Yang$^{\dagger}$, Lixing He$^{\dagger}$, Kaiwei Liu$^{\dagger}$, Zhenyu Yan$^{\dagger}$}
% \affiliation{\institution{$^{\dagger}$The Chinese University of Hong Kong\city{Hong Kong SAR}\country{China}} }

\author{\IEEEauthorblockN{Bufang Yang\IEEEauthorrefmark{2},
Lixing He\IEEEauthorrefmark{2}, Kaiwei Liu\IEEEauthorrefmark{2} and
Zhenyu Yan\IEEEauthorrefmark{2}}
\IEEEauthorblockA{
The Chinese University of Hong Kong, Hong Kong SAR, China\\
% Email: \IEEEauthorrefmark{1}author.one@add.on.net,
% \IEEEauthorrefmark{2}author.two@add.on.net,
% \IEEEauthorrefmark{3}author.three@add.on.net,
% \IEEEauthorrefmark{4}author.four@add.on.net
}
}

% \author{\IEEEauthorblockN{1\textsuperscript{st} Bufang Yang}
% \IEEEauthorblockA{\textit{The Chinese University of Hong Kong} \\
% % \textit{name of organization (of Aff.)}\\
% Hong Kong SAR, China \\
% bfyang@link.cuhk.edu.hk}
% \and
% \IEEEauthorblockN{2\textsuperscript{nd} Lixing He}
% \IEEEauthorblockA{\textit{The Chinese University of Hong Kong} \\
% % \textit{name of organization (of Aff.)}\\
% Hong Kong SAR, China \\
% 1155170464@link.cuhk.edu.hk}
% \and
% \IEEEauthorblockN{3\textsuperscript{rd} Kaiwei Liu}
% \IEEEauthorblockA{\textit{The Chinese University of Hong Kong} \\
% % \textit{name of organization (of Aff.)}\\
% Hong Kong SAR, China \\
% 1155189693@link.cuhk.edu.hk}
% \and
% \IEEEauthorblockN{4\textsuperscript{th} Zhenyu Yan}
% \IEEEauthorblockA{\textit{The Chinese University of Hong Kong} \\
% % \textit{name of organization (of Aff.)}\\
% Hong Kong SAR, China \\
% zyyan@ie.cuhk.edu.hk}

% \and
% \and

\maketitle

\begin{abstract}
% Statistics indicate that approximately 2.2 billion people worldwide suffer from visual impairment.
% Exploring artificial intelligence (AI) to support visually impaired (VI) individuals interests both companies and researchers.

% Visually impaired (VI) people refer to individuals who have a partial or total inability to visual perception.
% There are 2.2 billion people worldwide who suffer from visual impairments.
% Exploring artificial intelligence (AI) to support VI individuals has always attracted people's attention.

Individuals with visual impairments, encompassing both partial and total difficulties in visual perception, are referred to as visually impaired (VI) people. 
An estimated 2.2 billion individuals worldwide are affected by visual impairments. 
Recent advancements in multi-modal large language models (MLLMs) have showcased their extraordinary capabilities across various domains. 
% MLLMs can enable VI individuals remarkable visual understanding and reasoning capabilities.
It is desirable to help VI individuals with MLLMs' great capabilities of visual understanding and reasoning.
% However, images captured by VI individuals are significantly lower in quality than those taken by sighted individuals, which will lead MLLMs to generate unreliable responses.
However, it is challenging for VI people to use MLLMs due to the difficulties in capturing the desirable images to fulfill their daily requests. For example, the target object is not fully or partially placed in the image.
% Our preliminary findings indicate that current MLLMs can only generate general and impractical suggestions for VI individuals to reshoot their images. 
This paper explores how to leverage MLLMs for VI individuals to provide visual-question answers.
VIAssist can identify undesired images and provide detailed actions. 
Finally, VIAssist can provide reliable answers to users' queries based on the images.
% We design VIAssist, a MLLM that exhibits enhanced adaptability to the unique characteristics of queries from VI individuals.
% Qualitative and quantitative results show that VIAssist can generate more reliable responses, improving the usability of MLLMs for VI individuals.
Our results show that VIAssist provides +0.21 and +0.31 higher BERTScore and ROUGE scores than the baseline, respectively.
% Our result shows that VIAssist can generate more useful and reliable responses for queries posed by VI individuals.
% VIAssist can achieve +0.24 higher BERTScore and +0.31 ROUGE than the existing MLLMs.
% This paper first explores the limitations of existing MLLMs for VI queries including the images taken by VI and their questions. 
% Next, we design VIAssist, which can better adapt to the characteristics of VI queries and generate more useful and reliable responses. 

% This article aims to address these inquiries by focusing on the utilization of MLLMs to assist visually impaired (VI) individuals in their interactions with the physical environment.

% Our preliminary findings indicate that 1) a visually impaired person is difficult to take a clear and valid photo. 2) the current MLLMs provided by tech giants are not tailored for the disabled. Consequently, we design a system called VIAssist by tuning the model with a visually impaired-oriented dataset. Our result shows that the.
\end{abstract}


\section{Introduction}
Visually impaired (VI) people refer to individuals who have a partial or total inability to visual perception \cite{gurari2018vizwiz}.
According to the statistics from the World Health Organization (WHO), at least 2.2 billion people worldwide suffer from near or distance VI \cite{WHO}.
Recent years have seen growing interest in harnessing artificial intelligence (AI) technologies to support VI individuals \cite{Bemyai_2024,BeMyEyes_2024}.
% Exploring ways to leverage artificial intelligence (AI) technologies to assist VI individuals has attracted significant interest in recent years.

Many AI-based systems have been developed to enhance the quality of life for VI individuals \cite{kuriakose2023deepnavi,duh2020v}, e.g., obstacle detection and navigation. 
The exceptional performance and natural way of interaction exhibited by large language models (LLMs) and their multi-modal variants multi-modal large language models (MLLMs) have garnered increased attention \cite{liu2023llava,zhu2023minigpt,bai2023qwen}. 
VI individuals can query MLLMs like GPT-4V \cite{achiam2023gpt} to answer their daily concerns, such as ``What is the expiration date written on this medicine bottle?''
In the era of MLLMs, the remarkable visual understanding and reasoning capabilities of MLLMs can bring new life experiences to VI individuals.

However, it has been observed that the images captured by VI individuals are significantly lower in quality than those taken by sighted individuals\cite{gurari2018vizwiz}.
This is because VI individuals, due to their limited vision, cannot visually inspect captured images on their own. Consequently, such images are highly susceptible to poor quality, often capturing only parts of the intended target.
These low-quality images can lead the MLLMs to generate unreliable responses and decrease the experience of VI users.
A promising approach involves using MLLMs to guide VI individuals through a step-by-step process to capture high-quality images.
% is to add a prompt to the MLLMs, such as ``If the target can not be seen, please tell me how to retake the photo, e.g., how to adjust the shooting angle? ''.
However, our in-depth analysis shows that current MLLMs face challenges in providing effective and practical suggestions for VI users to re-shoot photos.

In this paper, we explore how to improve the usability of MLLMs for VI individuals.
% Our preliminary experiments show the current limitations of MLLMs in addressing VI-specific inquiries.
We design VIAssist, an MLLM tailored for enhanced adaptability to the unique inquiries of VI users.
For low-quality images, VIAssist can provide actionable and detailed suggestions for retaking photos. 
Upon capturing a high-quality photo, VIAssist is capable of producing reliable answers to queries from VI users.
We first collect an instruction dataset with questions and images tailored for VI individuals, alongside crafting aligned responses. 
Utilizing this dataset, our fine-tuned MLLM, VIAssist, showcases enhanced responsiveness to VI-specific queries. 
Both qualitative and quantitative results show that VIAssist can provide more reliable responses for VI users.

\section{Preliminary}

\subsection{Understanding VQA from VI individuals}
\subsubsection{Visual question answering (VQA)}
VQA has been a hot research topic in recent years \cite{kim2021vilt}.
The emergence of generative artificial intelligence, especially MLLMs \cite{liu2023llava,bai2023qwen}, has shifted VQA from closed-form to free-form, wherein it can generate text freely based on input images and questions.
% both before and after the era of foundation models. The basic VQA is close to image recognition where the expected output is a one-hot vector, except that there is a text description provided. VQA models require a text encoder so that the model is significantly heavier than the image recognition counterpart.
% The emergence of generative AI especially MLLMs shifts people's attention to more difficult free-form VQA, whose output is text wwithout any constraints.

\begin{figure}[t]
\begin{minipage}{0.47\columnwidth}
     \centering
\includegraphics[width=1\textwidth]{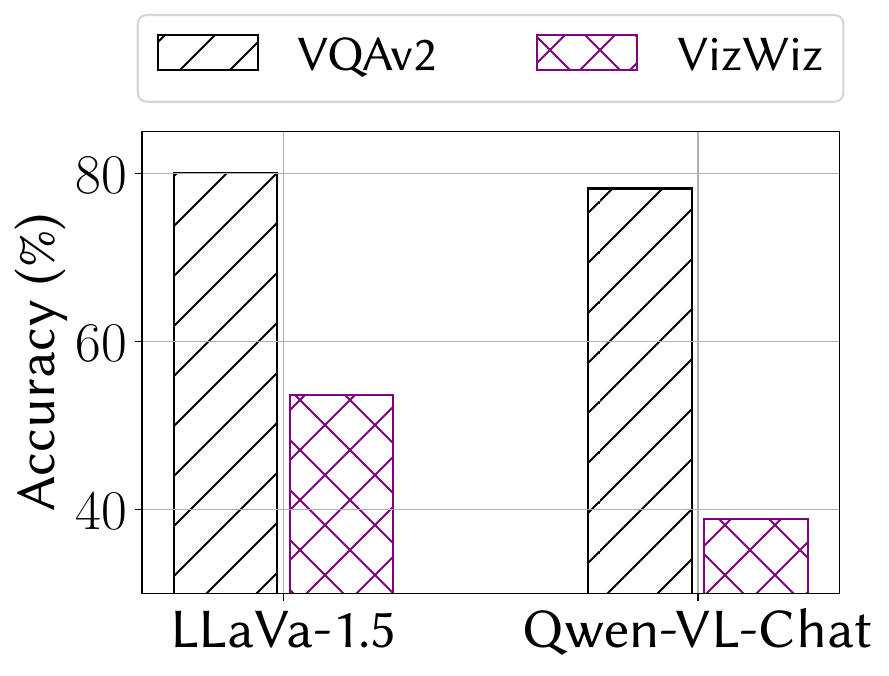}
        \caption{Performance of MLLMs on standard VQA and VI’s VQA datasets.
        }
        \label{fig:VQA_acc}
\end{minipage}
\hfill
   \begin{minipage}{0.47\columnwidth}
     \centering
\includegraphics[width=1\textwidth]{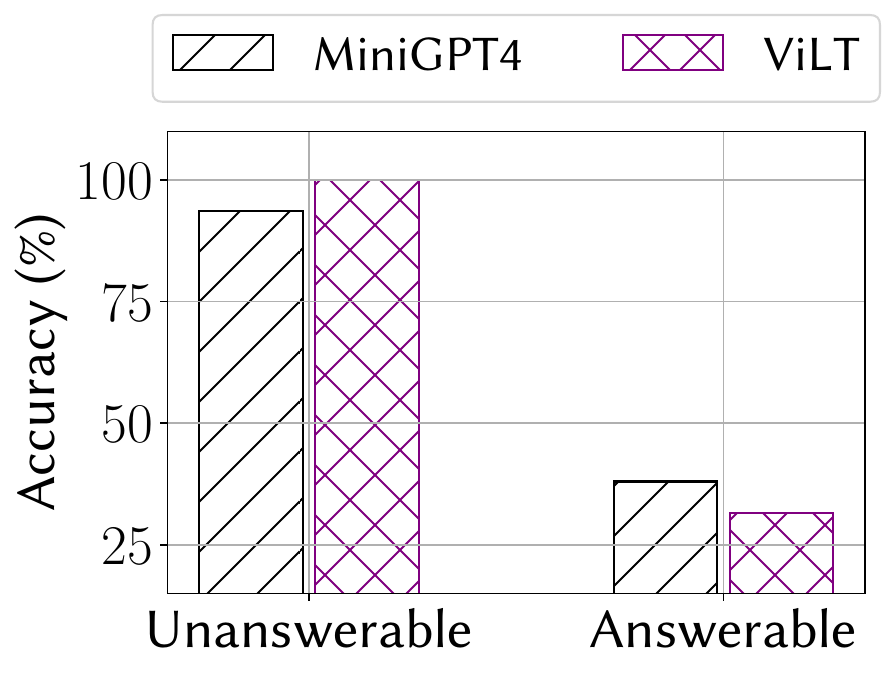}
    \caption{In-depth analysis of the MLLMs' performance on VI's VQA dataset.}
    % \vspace{-2em}
\label{fig:answerable}
\end{minipage}
% \vspace{-2.2em}
\end{figure}

\begin{figure}
  \centering
  \includegraphics[width=0.95\linewidth]{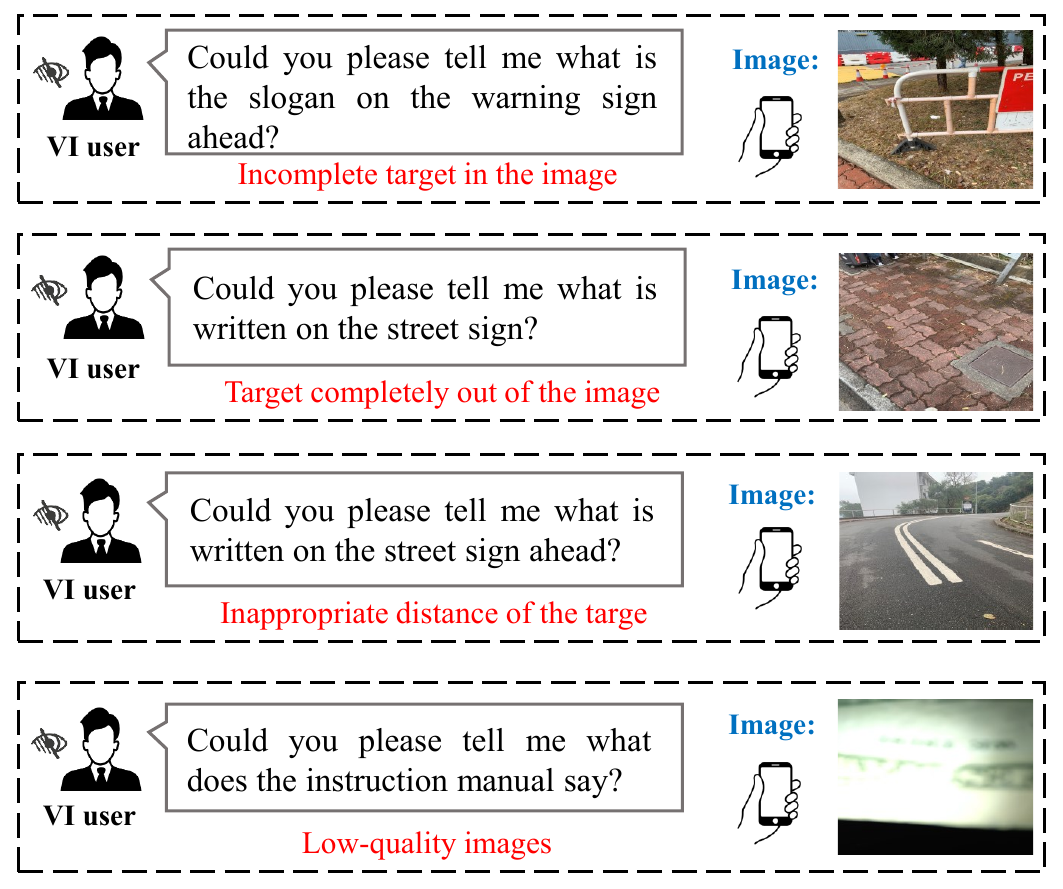}
  \caption{Examples of questions and corresponding photos from VI individuals.}
  % \vspace{-.5em}
  \label{fig:Examples}
  \vspace{-1em}
\end{figure}

\begin{figure*}[h]
  \centering
\includegraphics[width=1\linewidth]{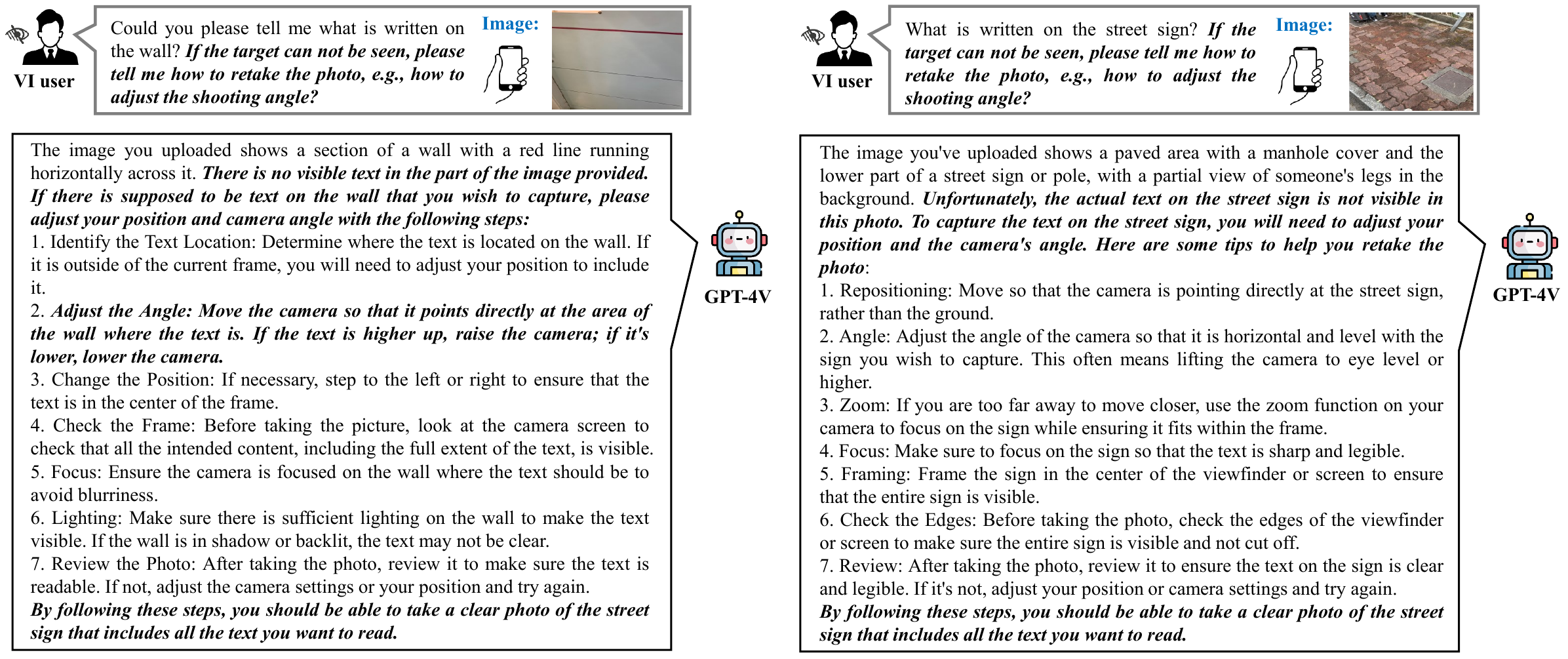}
  \caption{GPT-4V performance on 
  VI individual queries, where the target is completely out of the image.}
  % \vspace{-.5em}
  \label{fig:GPT-4V-Total}
  \vspace{-0.5em}
\end{figure*}

\begin{figure}
  \centering
\includegraphics[width=0.95\linewidth]{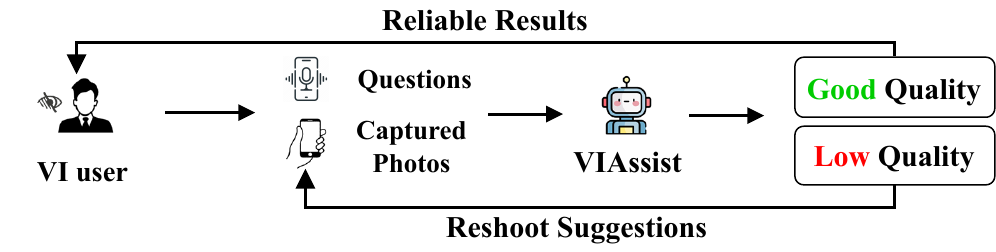}
  \caption{System overview of VIAssist.}
  % \vspace{-.5em}
  \label{fig:VIAssist_overview}
  \vspace{-1.2em}
\end{figure}

\begin{figure*}[h]
  \centering
\includegraphics[width=1\linewidth]{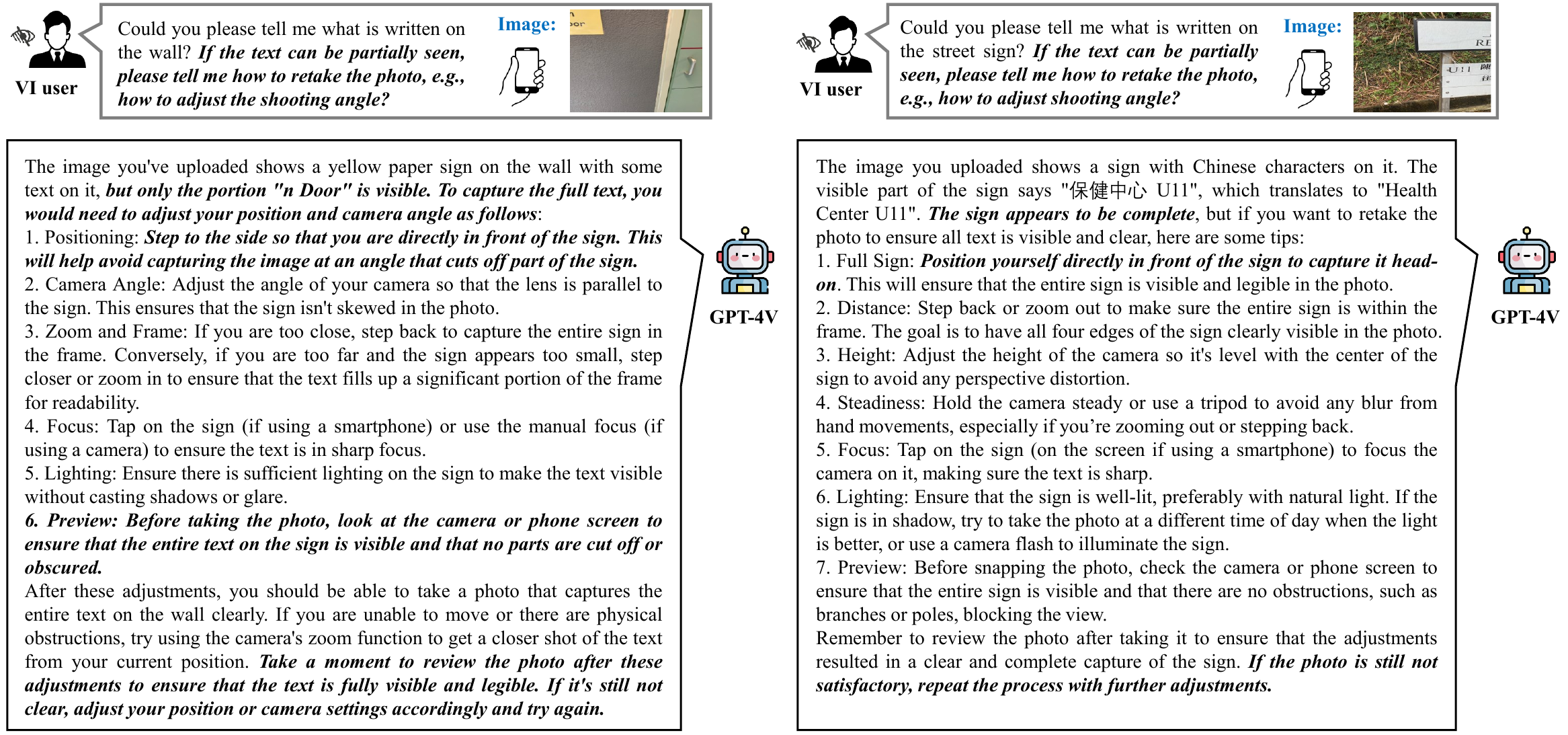}
  \caption{GPT-4V performance on 
  VI individual queries, where only a partial view of the target object mentioned in the question is captured in the image.}
  % \vspace{-.5em}
  \label{fig:GPT-4V-Partial}
  \vspace{-0.5em}
\end{figure*}

\subsubsection{VQA for VI individuals}
Building upon VQA, VQA for VI individuals (VI-VQA) is even more challenging.
Since the VI individual has poor eyesight, they cannot check the content of their captured image, resulting in a lower image quality. 
To the best of our knowledge, VizWiz \cite{gurari2018vizwiz} is the largest VQA dataset for blind people, consisting of 31,000 photos from blind people with ten crowd-sourced answers per visual question.
It is worth noting that among the photos taken by VI individuals, 28\% of them are classified as "Unanswerable," indicating the low quality of photos taken by blind people.
Therefore, providing VQA for VI individuals is a challenging task.

\subsubsection{Quantitative Analysis}

To analyze the difference between normal people and blind people, we test the performance of several popular MLLMs on two datasets: a standard VQA dataset VQAv2 \cite{goyal2017making} and a VI-VQA dataset, VizWiz. 
Figure~\ref{fig:VQA_acc} shows the performance of LLaVa-1.5 \cite{liu2023llava} and Qwen-VL-Chat \cite{bai2023qwen} on the two VQA datasets.
We can see that there is a significant accuracy gap between the two datasets:
in comparison to the VQAv2 dataset, the LLaVa-1.5 exhibits a decrease of 26.4\% accuracy on the VizWiz dataset, while the Qwen-VL-Chat demonstrates a reduction of 39.3\% accuracy.
It indicates that the images captured by VI individuals and their corresponding questions are more challenging than the standard VQA task. 
Even the recently proposed MLLMs struggle to handle them effectively.

% We follow the same prompt in [XXXX] to align the evaluation protocol of VizWiz. Since there are many ``Unanswerable'' images, we add ``if XXX, output unanswerable''.

% Our result replicates the same number in the paper [xxx], however, the number is biased and doesn't reflect the real performance. Specifically, LLM can give correct answers in most ``Answerable'' questions, while makes huge mistakes in ``Unanswerable'' questions. This discovery clearly states that LLMs trained from large-scale dataset and LLMs' benchmark omit the need of blind community.

Next, we further analyze the VizWiz dataset in depth by whether the question is ``Answerable'' or not (2934 ``Answerable'' questions and 1385 ``Unanswerable'' questions). 
We test two models including one MLLM, MiniGPT4 \cite{zhu2023minigpt}, and one traditional VQA model, ViLT \cite{kim2021vilt}.
Figure~\ref{fig:answerable} shows that both MiniGPT4 and ViLT models accurately predict ``Unanswerable'' questions.
However, they achieve poor accuracy for ``Answerable'' questions, lower than 40\%.
Through a more detailed examination of the VizWiz dataset, we find that even for the "Answerable" questions, the corresponding images exhibit relatively lower quality, e.g., only part of the target is captured in the image.

In summary, we find that the reason for MLLMs' low accuracy lies in two aspects: 1) the poor quality of captured images by VI individuals, and 2) many photos taken by VI individuals are not ``Answerable''. We will investigate the features of such gaps and the reasons behind them in the next subsection.

\subsubsection{Key features of VI queries}
\label{Key features}

In this subsection, we outline the key characteristics of queries from VI individuals. 
These features are crucial for understanding the unique needs and challenges faced by VI individuals when interacting with MLLMs. 
% We aim to facilitate the growth of MLLMs that are more accessible and user-friendly for VI individuals.
We summarize the key features of VI queries as follows:
\begin{itemize}[leftmargin=*]

\item  \textbf{Incomplete target in the image.}
Only a partial view of the target object mentioned in the question is captured in the image, such as the left part, right part, upper part, or bottom part.

\item  \textbf{Target completely out of the image.}
The target object in the question is entirely absent from the image.

\item  \textbf{Inappropriate distance of the target.}
The distance of the target object in the image is inappropriate, such as being either too far or too close.

\item  \textbf{Low-quality images.}
The captured images suffer from blurriness, complete darkness, or low light conditions.

\item  \textbf{Nonsense or irrelevant questions.}
The content of the pictures taken is completely unrelated to the question posed by the VI individual.

\end{itemize}

Figure~\ref{fig:Examples} shows some examples of queries and captured images from VI individuals. 
These types of queries often arise, and they are inherently challenging or even impossible to answer, even for humans. Given that MLLMs are trained and fine-tuned based on human instruction, it is unsurprising that MLLMs may encounter difficulties in providing responses in such cases.

\subsection{Analysis of MLLMs for VI Queries}
We collect a set of low-quality images from the real world and request GPT-4V to generate responses to our inquiries.
For GPT-4V to give suggestions for reshooting low-quality images, we set the prompt of GPT-4V as ``\texttt{<Question>}. If the target can not be seen, please tell me how to retake the photo, e.g., how to adjust the shooting angle?''.
Figure~\ref{fig:GPT-4V-Total} and Figure~\ref{fig:GPT-4V-Partial} show the GPT-4V performance on these low-quality images.
We observe that GPT-4V's queries for VI individuals exhibit the following characteristics:

% \noindent\textbf{General and high-level guidance.}
\subsubsection{General and high-level guidance}
Figure~\ref{fig:GPT-4V-Total} shows the scenario where the target, such as the text on the street sign, is entirely absent in the captured image.
% To enhance GPT-4V's ability to generate instructions on adjusting the shooting angle and retaking a photo, we incorporate an additional prompt:``If the target can not be seen, please tell me how to retake the photo, e.g., how to adjust the shooting angle? ''
We can see that GPT-4V has the ability to identify the targets that are not visible in the given photo. 
However, the generated suggestions for retaking photos remain limited to general and high-level guidance on how to capture a picture with higher quality.
The response generated by GPT-4V lacks specific guidance for VI individuals on adjusting the shooting angle, including instructions on how to adjust the shooting direction and which direction to move in.
This limitation restricts its practical utility in assistive systems designed for blind people.

Figure~\ref{fig:GPT-4V-Partial} shows an example where the target is partially absent in the captured image.
The result shows that GPT-4V can successfully detect the target's absence. 
However, when we add the prompt: ``If the text can be partially seen, please tell me how to retake the photo, e.g., how to adjust the shooting angle?'', GPT-4V only provides general and high-level guidance, failing to provide VI people helpful suggestions.

% \noindent\textbf{Impractical and ineffective suggestions for VI individuals.}
\subsubsection{Impractical and ineffective suggestions for VI individuals}
Figure~\ref{fig:GPT-4V-Partial} also shows that GPT-4V even suggests VI individuals look at the photo to ensure the target's visibility, which is an impractical suggestion for VI individuals.

In summary, current MLLMs like GPT-4V have the ability to detect low-quality images and identify whether the target is partially or completely absent. However, their responses are limited to high-level and general suggestions on adjusting the shooting and even impractical suggestions for VI individuals.
These MLLMs can not provide a detailed or step-by-step guide for capturing high-quality images that would enable VI individuals to address their specific inquiries.
The main challenge lies in enhancing MLLMs to generate detailed and practical suggestions for reshooting when the quality of the images provided by VI individuals is low.

\section{VIAssist}

In this paper, we design VIAssist, which can adapt to the characteristics of VI individual queries, and generate reliable responses.
Figure~\ref{fig:VIAssist_overview} shows the system overview of VIAssist.
For low-quality images, VIAssist provides actionable and detailed advice on retaking photos. 
VIAssist can generate reliable answers to queries from VI users based on the given images.

\subsection{Instruction Dataset}
To enhance the MLLMs to generate reliable and practical responses for VI individuals, we need to collect an instruction dataset for fine-tuning.
The format of each sample in our dataset is as follows: $<$ Question, Image, Response $>$.

\noindent\textbf{Image Collection.}
According to our analysis of VI individual queries in \S~\ref{Key features}, we collect the following types of images and questions:
1) High-quality images and questions,
2) Target completely out of the image,
3) Incomplete target in the image,
4) Inappropriate distance of the target,
5) Low-quality image,
6) Nonsense or irrelevant questions.

Since VizWiz dataset \cite{gurari2018vizwiz} lacks types 2), 3), and 4) of images and questions, we collect a real-world dataset at various locations such as campus roads, shopping malls, streets, and indoor corridors.

% \begin{itemize}[leftmargin=*]
% \item  \textbf{High quality images and questions.}
% Both the captured images and the posed questions are clear, allowing for answers based on the information present in the images.

% \item  \textbf{Target completely out of the image.}
% The target object in the question is entirely absent from the image.

% \item  \textbf{Incomplete target in the image.}
% Only a partial view of the target object mentioned in the question is captured in the image, such as the left part, right part, upper part, or bottom part.

% \item  \textbf{Inappropriate distance of the target.}
% The distance of the target object in the image is inappropriate, such as being either too far or too close.

% \item  \textbf{Low-quality image.}
% The captured images suffer from blurriness, complete darkness, or low light conditions.

% \item  \textbf{Nonsense or irrelevant questions.}

% \end{itemize}

% \noindent\textbf{Image Collection.}
% Our image dataset contains two parts: the public dataset VizWiz and our self-collected real-world dataset.
% VizWiz dataset contains the real images taken by visual impaired patients and their spoken questions about the images.
% Besides, we will collect another real-world dataset using mobile phones.
% \textcolor{blue}{The reason why we collect a real-world dataset is that VizWiz lacks type II and type III images and questions.}
% We will gather images from various locations such as campus roads, shopping malls, streets, indoor corridors, and more.

\noindent\textbf{Question Collection.}
% For images from the VizWiz dataset, we will directly utilize the original questions provided.
For each collected image, we first manually create questions.
% Given the varying questioning styles among different users, 
We also employ GPT-4V to rewrite and expand the diversity of created questions.

\noindent\textbf{Response Collection.}
We recruited annotators to manually write responses for each image and question we collected.
The annotated response contains descriptions and suggestions. 
The description part refers to GPT-4V's answer. 
In the case of images captured with good quality, the response will include the statement ``the quality of this image is good.'' However, for images with poor shooting quality, such as only part of the target can be seen in the image, a description of the poor quality and suggestions for how to adjust the shooting will be added to the response.

Currently, we have gathered 50 images for each type of VI query.
There are 300 $<$ Question, Image, Response $>$ pairs in our instruction dataset in total.
Additionally, we have collected a separate set of 50 images for evaluation. 
These images are captured at various locations, distinct from the ones used in the instruction dataset.

% In the future, we plan to design a questionnaire that contains crucial factors for evaluation, e.g., is relevant information found in the image?
% Such assisted questionnaires can reduce labeling bias among annotators and efforts (structured information?).

\subsection{Model Training}
\noindent\textbf{Model Architecture.}
The model architecture of VIAssist is based on the open-source MLLM, LLaVA \cite{liu2023llava}.
It contains a pre-trained visual encoder and an LLM.
The visual features are converted to the word embedding space through a projection layer.
We use the frozen CLIP ViT-L/14 \cite{radford2021learning} as the visual encoder, and use Vicuna-7B \cite{chiang2023vicuna} as the LLM.

\noindent\textbf{Instruction Tuning.}
Since MLLM fine-tuning with full parameters requires a huge amount of data and computation resources, we use LoRA \cite{hu2021lora} technique for
parameter-efficient fine-tuning.
We add LoRA parameters to the projection layers within the attention layers of the LLM.
During training, only the feature projectors and LoRA parameters are updated, while the remaining parameters are kept frozen.

\subsection{VIAssist Performance}
\begin{figure*}[h]
  \centering
\includegraphics[width=0.9\linewidth]{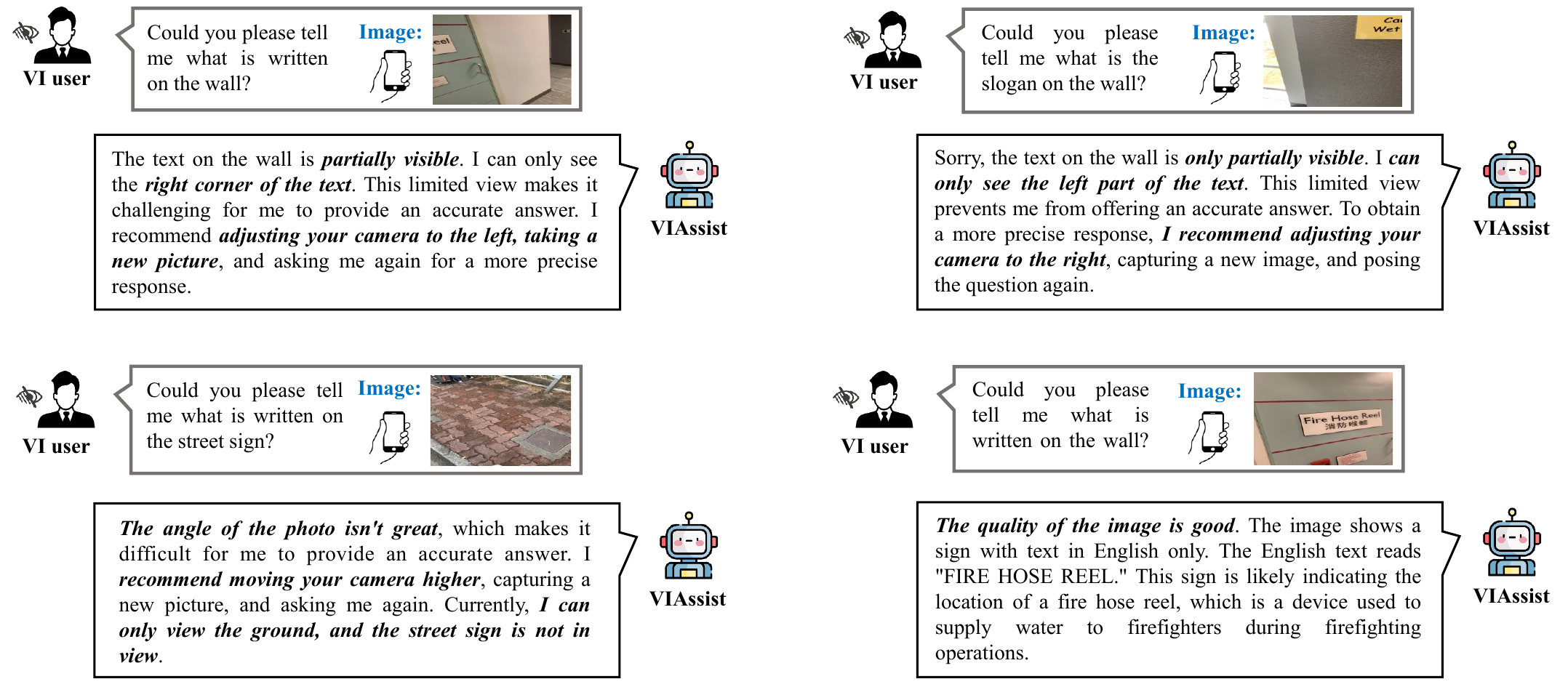}
  \caption{VIAssist performance on VI individual queries. It can assess the image quality, generate specific suggestions for adjusting the shot, and contain fewer irrelevant responses.}
  % \vspace{-.5em}
  \label{fig:VIAssist}
  \vspace{-0.5em}
\end{figure*}

% \subsubsection{Qualitative Results}
\noindent\textbf{Qualitative Results.}
Figure~\ref{fig:VIAssist} shows the performance of VIAssist.
The results show that VIAssist's response exhibits three distinct advantages over GPT-4V.
Firstly, it can assess the image quality and provide explanations for poor quality.
Secondly, it offers more \textbf{detailed} and \textbf{actionable} suggestions for adjusting the shot, such as moving the camera to the left and taking another shot.
In addition, it can generate fewer irrelevant or nonsensical responses.
Overall, VIAssist can generate more reliable responses for VI individuals.

\begin{figure}
\begin{minipage}[t]{0.6\columnwidth}
     \centering
\includegraphics[width=1\textwidth]{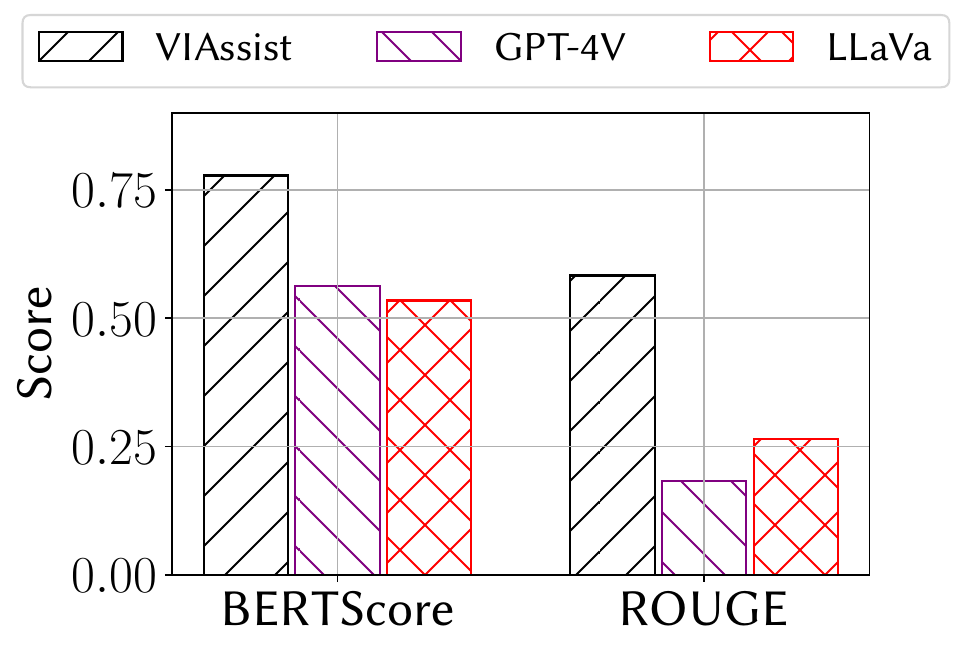}
        \caption{Quantitative results of VIAssist and other MLLMs.
        }
        \label{fig:results}
\end{minipage}
\hfill
   \begin{minipage}[t]{0.37\columnwidth}
     \centering
\includegraphics[width=1\textwidth]{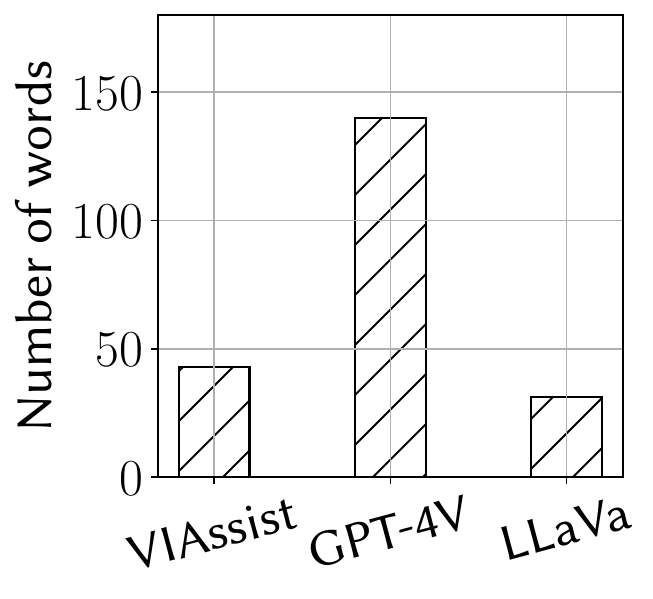}
    \caption{The number of words in MLLMs' responses.}
    \vspace{-1em}
\label{fig:results_words}
\end{minipage}
% \vspace{-2em}
\end{figure}

% \subsubsection{Quantitative Results}
\noindent\textbf{Quantitative Results.}
Following \cite{zhao2024vialm}, we use BERTScore and ROUGE as two evaluation metrics to quantify the performance of VIAssist and existing MLLMs in terms of response quality.
BERTScore \cite{zhang2019bertscore} can be used to assess the semantic similarity between two texts.
ROUGE \cite{lin2004rouge} can be used to assess the degree of token overlap between two texts.
For LLaVA, we use LLaVA-v1.6-34B version.
For GPT-4V, we set the prompt as ``\texttt{<Question>}. If the target can not be seen, please tell me how to retake the photo, e.g., how to adjust the shooting angle?''.
We test the performance of LLaVA and GPT-4V under the zero-shot setting.

Figure~\ref{fig:results} shows that VIAssist achieves +0.21 higher BERTScore and +0.31 higher ROUGE than the best baseline model.
It demonstrates that VIAssist can generate more reliable responses to queries posed by VI individuals.
Figure~\ref{fig:results_words} shows the number of words in MLLMs’ responses.
We can see that GPT-4V generates the longest responses, likely due to its extensive incorporation of world knowledge, resulting in diverse content. 
However, as shown in Figure~\ref{fig:GPT-4V-Total} and Figure~\ref{fig:GPT-4V-Partial}, GPT-4V's responses often include a significant number of irrelevant or nonsensical suggestions. 
These suggestions are general and impractical for VI individuals.
In contrast, VIAssist's responses exhibit higher semantic scores, moderate word count, and complexity, thereby delivering an enhanced QA experience for VI individuals.

% \begin{figure}
%   \centering
% \includegraphics[width=0.6\linewidth]{VIAssist/results.pdf}
%   \caption{Quantitative performance of VIAssist and other MLLMs.}
%   % \vspace{-.5em}
%   \label{fig:results}
%   % \vspace{-1em}
% \end{figure}

\section{FUTURE DIRECTIONS}
\noindent\textbf{Instruction Dataset.}
The instruction dataset collected in this study is still limited, which occasionally leads VIAssist to provide inaccurate reshoot suggestions for new data.
In the future, we will enrich our image dataset with greater diversity, including photos taken by real VI individuals, to enhance the practicality and generalization of VIAssist.

\noindent\textbf{Automatic Reshooting.}
We also plan to enhance the usability of VIAssist. 
A potential approach involves refining VIAssist's output (guidance) to enable automatic camera adjustments, such as zooming in or out, eliminating the need for users to comprehend the guidance and manually retake photos

\noindent\textbf{Real-time and Efficiency.}
According to Be My AI \cite{Bemyai_2024}, which is the popular AI-based VI assistant app, the response time on average is still 4 minutes, which is not satisfying if the VI user wants to actively use it. 
Increasing computing resources is a direct approach to addressing this issue. 
However, this poses challenges for budget-constrained service providers like Be My Eyes \cite{BeMyEyes_2024}, requiring significant effort to generate revenue.
% Another promising direction is Edge LLM.
% Qualcomm announced that the snapdragon 8 Gen 3 will have strong generative AI capability (up to 20 tokens/second).
% Besides the above approaches that focus on better platforms,
In addition, VI users can also capture videos to gather more information about their surroundings, which presents substantial challenges in terms of edge computing costs, network bandwidth, and MLLMs cloud services.
Several neural network (NN) inference optimization and video streaming techniques can be considered to enhance the efficiency of the VI assist system, including edge-cloud collaboration \cite{yang2023edgefm}, NN quantization \cite{ma2024era}, and the efficient NN architectures \cite{dao2022flashattention}.

\noindent\textbf{Dedicated Prompt Engineering.}
This work has not yet meticulously designed prompts for GPT-4V. Nonetheless, it is worth noting that prompt engineering \cite{liu2023pre} can significantly affect the performance of MLLMs. 
Future research could explore more effective prompts to enhance GPT-4V's responses to inquiries from VI users.

\noindent\textbf{Other Modalities.}
% Using out-of-shelf devices like smartphones is the easiest way to spread the technique to a vast group of people. 
% In contrast, there is no need to only work on smartphones for blind people since they are possible to purchase dedicated devices to help them.
When the target is completely absent from the photo, it poses a challenge for MLLMs to provide effective reshooting suggestions since no useful information is captured in the image.
In contrast to sighted individuals, blind individuals commonly carry additional assistive devices, such as guide crutches, in addition to mobile phones.
% AiSee \cite{National_University_of_Singapore_2024} is one example designed by the National University of Singapore. 
% It is a wearable device that consists of 1) image processing 2) natural question answers 3) bone conduction feedback. 
% In other words, AiSee simplifies the task so it is not necessary to incorporate powerful LLMs. 
One open question is: can we leverage additional sensors and modalities, such as wireless signals, audio, and Electroencephalogram (EEG) \cite{yang2022novel,yang2021single}, to improve the performance of MLLMs?

\noindent\textbf{Other Types of Impaired People.}
Beyond aiding VI users, LLMs can assist with other prevalent disabilities. 
% Just as with vision, LLMs can serve as a central point for integrating various modalities. 
For instance, individuals who are deaf or hard of hearing may benefit from LLMs to interpret ambient sounds.
% Except visually impaired, there are other common disabilities that LLMs can help. 
% Similar to vision, LLMs can act as a hub to incorporate other modalities. For example, people who are deaf and hard of hearing would like to understand the surrounding sound. 
However, previous work ProtoSound \cite{jain2022protosound} still relied on a relatively simple classification network, which leaves space for improvement.

\section{Related work}
Prior to the advent of MLLMs, researchers also explored the utilization of diverse AI models to assist VI individuals.
Lin \textit{et al.} \cite{lin2019deep} propose a multi-modal VI assist system that leverages data from RGBD cameras and earphones to aid blind individuals in navigating and understanding environments.
DeepNAVI \cite{kuriakose2023deepnavi} runs multiple AI models simultaneously to give VI users walking instructions. 
As for V-eye \cite{duh2020v}, combining global localization and image segmentation, the users can be given precise location information.
In summary, these works are ad-hoc designed and provide only partial information needed by VI users. Given the exceptional visual understanding and reasoning capabilities of MLLMs, exploring the application of MLLMs for VI users holds considerable promise.

% In summary, all those works are ad-hoc designed and only give partial information needed by blind people. 
% Since MLLMs already present a great ability for generalization, it is more promising to work on general assistants. 

% In the era of MLLMs, the exceptional visual understanding and reasoning capabilities they possess have the potential to revolutionize the life experiences of VI individuals.
% However, 
Recent studies on MLLMs \cite{liu2023llava,bai2023qwen} mainly cater to users with unimpaired vision, offering scant consideration for individuals with visual impairments.
Liu \textit{et al.} \cite{liu2023open} incorporates SAM \cite{kirillov2023segment} and image captioning to build a system to help blind people. 
Be My Eyes \cite{BeMyEyes_2024} is a free mobile application designed to enhance accessibility for individuals who are blind or have low vision. 
The application has released its new function: Be My AI \cite{Bemyai_2024}, which utilizes AI models to guide users instead of volunteers. 
Compared to the original Be My Eyes, it provides a 90\%+ successful rate with a one-third response time (4 minutes on average). In other words, only 10\% of calls will be handled by volunteers. However, the response time is still too long to be applied in the real world.
VIALM \cite{zhao2024vialm} is the first benchmark to test the performance of MLLMs for queries from VI individuals. 
However, it primarily focuses on evaluating the success rate of fulfilling requests from VI individuals, overlooking the issues of low-image quality conditions.

\section{Conclusion}
This paper aims to enhance the usability of MLLMs for VI individuals.
Our preliminary experiments show the limitations of current MLLMs when generating responses for inquiries from VI individuals.
In this paper, we design VIAssist, a MLLM that exhibits enhanced adaptability to the unique characteristics of VI queries.
For low-quality images, VIAssist provides actionable and detailed advice on retaking photos. Once a high-quality photo is captured, VIAssist can deliver reliable answers to queries from VI users.
% We first curated an instruction dataset with questions and images tailored for VI individuals, alongside crafting aligned responses. 
% Utilizing this dataset, our fine-tuned MLLM, VIAssist, showcases enhanced responsiveness to VI-specific queries. 
Both qualitative and quantitative results show that VIAssist can generate more reliable and practical responses for VI users.
% In this paper, we summarize the characteristics of VI queries and analyze the limitations of current MLLMs when generating responses for inquiries from VI individuals.
% To address these challenges, we design VIAssist, which demonstrates improved adaptability to the unique characteristics of queries from VI individuals.
% Our qualitative and quantitative results show that VIAssist can generate more useful and reliable responses for VI individuals, which enhances the usability of MLLMs.

\bibliographystyle{ieeetr}
\bibliography{main}

\end{document}